\documentclass{article}

 \usepackage[final,nonatbib]{neurips_2019}

\usepackage[utf8]{inputenc} % allow utf-8 input
\usepackage[T1]{fontenc}    % use 8-bit T1 fonts
\usepackage{hyperref}       % hyperlinks
\usepackage{url}            % simple URL typesetting
\usepackage{booktabs}       % professional-quality tables
\usepackage{amsfonts}       % blackboard math symbols
\usepackage{nicefrac}       % compact symbols for 1/2, etc.
\usepackage{microtype}      % microtypography
\usepackage{amsmath}
\usepackage{makecell}

\usepackage[dvipsnames]{xcolor}
\usepackage{tikz}
\usetikzlibrary{shapes,arrows,calc}
\tikzstyle{block2} = [rectangle, draw, fill=Gray!10, text width=10em, text centered, rounded corners, minimum height=2em]
\tikzstyle{block3} = [rectangle, draw, fill=Gray!0, text width=10em, text centered, rounded corners, minimum height=2em]
\tikzstyle{block4} = [rectangle, draw, fill=Gray!0, text width=13em, text centered, rounded corners, minimum height=2em]
\tikzstyle{line} = [draw, -latex']
\tikzstyle{doc}=[%
draw,
thick,
align=center,
color=black,
shape=document,
minimum width=1cm,
minimum height=1cm,
shape=document,
inner sep=2ex,
]

\newcommand{\arash}[1]{}
\renewcommand{\arash}[1]{{\color{red} [Arash: {#1}]}}
\newcommand{\mrinal}[1]{}
\renewcommand{\mrinal}[1]{{\color{red} [Mrinal: {#1}]}}
\newcommand{\sonal}[1]{}
\renewcommand{\sonal}[1]{{\color{red} [Sonal: {#1}]}}

\title{ Improving Robustness of Task Oriented Dialog Systems}

\author{
  Arash Einolghozati\\
  Facebook Assistant\\
  \texttt{arashe@fb.com} \\
   \And
   Sonal Gupta \\
  Facebook Assistant\\
   \texttt{sonalgupta@fb.com} \\
   \AND
   Mrinal Mohit \\
  Facebook Assistant\\
   \texttt{mrinalmohit@fb.com} \\
    \And
   Rushin Shah\thanks{Work done while at Facebook} \\
     Google\\
   \texttt{rushinshah@google.com} 
}

\begin{document}

\maketitle

\begin{abstract}
  Task oriented language understanding in dialog systems is often modeled using intents (task of a query) and slots (parameters for that task). Intent detection and slot tagging are, in turn, modeled using sentence classification and word tagging techniques respectively. Similar to adversarial attack problems with computer vision models discussed in existing literature, these intent-slot tagging models are often over-sensitive to small variations in input -- predicting different and often incorrect labels when small changes are made to a query, thus reducing their accuracy and reliability. However, evaluating a model's robustness to these changes is harder for language since words are discrete and an automated change (e.g. adding `noise') to a query sometimes changes the meaning and thus labels of a query. In this paper, we first describe how to create an adversarial test set to measure the robustness of these models. Furthermore, we introduce and adapt adversarial training methods as well as data augmentation using back-translation to mitigate these issues. Our  experiments  show that both techniques improve the robustness of the system substantially and can be combined to yield the best results.
\end{abstract}

\section{Introduction}

In computer vision, it is well known that otherwise competitive models can be "fooled" by adding intentional noise to the input images~\cite{Szegedy_intriguing_2014, Papernot2017Blackbox}. Such changes, imperceptible to the human eye, can cause the model to reverse its initial correct decision on the original input. This has also been studied for Automatic Speech Recognition (ASR) by including hidden commands~\cite{hidden_commands} in the voice input. Devising such adversarial examples for machine learning algorithms, in particular for neural networks, along with defense mechanisms against them, has been of recent interest~\cite{madry2018towards}. The lack of smoothness of the decision boundaries~\cite{Goodfellow2014Explain} and reliance on weakly correlated features that do not generalize~\cite{robustness2018odds} seem to be the main reasons for confident but incorrect predictions for instances that are far from the training data manifold. Among the most successful techniques to increase resistance to such attacks is perturbing the training data and enforcing the output to remain the same~\cite{Goodfellow2014Explain,logit_pairing_2018}. This is expected to improve the smoothing of the decision boundaries close to the training data but may not help with points that are far from them.
% Miyato et al introduce virtual adversarial training defense against the fast gradient attacks~\cite{Goodfellow}.

There has been recent interest in studying this adversarial attack phenomenon for natural language processing tasks, but that is harder than vision problems for at least two reasons: 1) textual input is discrete, and 2) adding noise may completely change a sentence's meaning or even make it meaningless. Although there are various works that devise adversarial examples in the NLP domain, defense mechanisms have been rare. \cite{Miyato_VAT_NLP} applied perturbation to the continuous word embeddings instead of the discrete tokens. This has been shown~\cite{adversarial_pos} to act as a regularizer that increases the model performance on the clean dataset but the perturbed inputs are not true adversarial examples, as they do not correspond to any input text and it cannot be tested whether they are perceptible to humans or not. 

Unrestricted adversarial examples~\cite{unrestricted_goodfellow} lift the constraint on the size of added perturbation and as such can be harder to defend against. Recently, Generative Adversarial Networks (GANs) alongside an auxiliary classifier have been proposed to generate adversarial examples for each label class. In the context of natural languages, use of seq2seq models~\cite{seq2seq}
seems to be a natural way of perturbing an input example~\cite{luke_adversarial}. Such perturbations, that practically paraphrase the original sentence, lie somewhere between the two methods described above. On one hand, the decoder is not constrained to be in a norm ball from the input and, on the other hand, the output is strongly conditioned on the input and hence, not unrestricted.
 
Current NLP work on input perturbations and defense against them has mainly focused on sentence classification. In this paper, we examine a harder task: joint intent detection (sentence classification) and slot tagging (sequence word tagging) for task oriented dialog, which has been of recent interest \cite{HakkaniTur2016} due to the ubiquity of commercial conversational AI systems. 

In the task and data described in Section~\ref{sec:task}, we observe that exchanging a word with its synonym, as well as changing the structural order of a query can flip the model prediction. Table~\ref{tab:example} shows a few such sentence pairs for which the model prediction is different. Motivated by this, in this paper, we focus on analyzing the model robustness against two types of untargeted (that is, we do not target a particular perturbed label) perturbations: paraphrasing and random noise. In order to evaluate the defense mechanisms, we discuss how one can create an adversarial test set focusing on these two types of perturbations in the setting of joint sentence classification and sequence word tagging.

%\sonal{can be condensed} We first investigate our clean dataset for sentences that seem to be perturbations of each other, but for which the model makes different decisions. Then, we explore methods to generate such perturbations. We define a semantic frame as the combination of the sentence label and the sorted list of contiguous word labels. Upon analyzing the model output for sentences belonging to the same frame, we observed that exchanging a word with its synonym, as well as changing the structural order of a query can flip the model prediction. Table~\ref{tab:example} shows a few such sentence pairs for which the model prediction is different. We observe that after replacing slots with their labels, sentences having the same frame can be considered as loose paraphrases of each other.

Our contributions are: 1. Analyzing the robustness of the joint task of sentence classification and sequence word tagging through generating diverse untargeted adversarial examples using back-translation and noisy autoencoder, and 2. Two techniques to improve upon a model's robustness   -- data augmentation using back-translation, and adversarial logit pairing loss. Data augmentation using back-translation was earlier proposed as a defense mechanism for a sentence classification task~\cite{luke_adversarial}; we extend it to sequence word tagging. We investigate using different types of machine translation systems, as well as different auxiliary languages, for both test set generation and data augmentation. Logit pairing was proposed for improving the robustness in the image classification setting with norm ball attacks~\cite{logit_pairing_2018}; we extend it to the NLP context.  We show that combining the two techniques gives the best results. 
 
%The rest of the paper is organized as follows: In Section~\ref{sec:task}, we introduce the task, the data, and the base model as well as the robustness measure. In Section~\ref{sec:attacks}, we introduce two methods of devising adversarial examples for the aforementioned model and in Section~/ref{sec:defense}, we discuss the data augmentation and model-based mechanisms to improve the adversarial test accuracy and the robustness measure. 

\begin{table*}[hbt!]
\centering
\small
\begin{tabular}{lll}
  Labels  & Correct Prediction & Wrong Prediction \\
  \hline
  (weather/check\_sunset)& When is dusk & When is the sunset \\
  (alarm/modify\_alarm, datetime) & Adjust alarm to 9am & Extend alarm by 20 minutes \\
  (weather/find, datetime, location)&What's the high today in Sydney & What's the high in Portland today  \\
\end{tabular}
\caption{Pairs of sentences whose oracle labels are same but the base model predictions are different. Note that small changes to an sentence results in different model predictions.}
\label{tab:example}
\end{table*}

\section{Task and Data}
\label{sec:task}

\tikzstyle{block} = [text width=15em, text centered]

\begin{figure}
    \centering
    \small
    \begin{tikzpicture}
    \node [block] (intent) {weather/find};
    \node [block, below of=intent, node distance=2.6em] (sentence) {What's the weather in Sydney today};
    \draw[draw=black]  ($(sentence.west)+(0,0.3)$) -- ($(sentence.east)+(0,0.3)$);
    \node [block, below of=sentence, node distance=2.5em, xshift=3em] (slot1) {location};
    \node [block, below of=sentence, node distance=2.5em, xshift=6.8em] (slot2) {datetime};
    \path [line] (sentence.north) -| (intent.south);
    \path [line] ($(sentence.south)+(0.95,0)$) -| (slot1.north);
    \path [line] ($(sentence.south)+(2.15,0)$) -| (slot2.north);
    \end{tikzpicture}
    \caption{Example sentence and its annotation}
    \label{fig:intent_slot_tagging}
\end{figure}
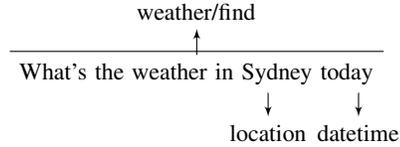

In conversational AI, the language understanding task typically consists of classifying the intent of a sentence and tagging the corresponding slots. For example, a query like \textit{What's the weather in Sydney today} could be annotated as a \textit{weather/find} intent, with \textit{Sydney} and \textit{today} being \textit{location} and \textit{datetime} slots, respectively. This predicted intent then informs which API to call to answer the query and the predicted slots inform the arguments for the call. See Fig.~\ref{fig:intent_slot_tagging}.  Slot tagging is arguably harder compared to intent classification since the spans need to align as well.

We use the data provided by~\cite{sebastianMultilingual}, which consists of task-oriented queries in weather and alarm domains. The data contains 25k training, 3k evaluation and 7k test queries with 11 intents and 7 slots.
We conflate and use a common set of labels for the two domains. Since there is no ambiguous slot or intent in the domains, unlike~\cite{onenet}, we do not need to train a domain classifier, neither jointly nor at the beginning of the pipeline. If a query is not supported by the system but it is unambiguously part of the alarm or weather domains, they are marked as \textit{alarm/unsupported} and \textit{weather/unsupported} respectively.

% \begin{table*}[t!]
% \centering
% \small
% \begin{tabular}{lll}
%   Frame  & Correct  & Wrong \\
%   \hline
%   (weather/check\_sunset,)& When is dusk & When is the sunset \\
%   (alarm/modify\_alarm, datetime) & Adjust alarm to 9am & Extend alarm by 20 minutes \\
%   (weather/find, datetime, location)&What's the high today in Sydney & What's the high for Portland today  \\
% \end{tabular}
% \caption{Pairs of sentences whose oracle labels are same but model's predictions are different.}
% \label{tab:example}
% \end{table*}

\section {Robustness Evaluation}
\label{sec:attacks}

 To evaluate model robustness, we devise a test set consisting of ‘adversarial’ examples, i.e, perturbed examples that can potentially change the base model's prediction. These could stem from paraphrasing a sentence, e.g., lexical and syntactical changes. We use two approaches described in literature: back-translation and noisy sequence autoencoder. Note that these examples resemble black-box attacks but are not intentionally designed to fool the system and hence, we use the term 'adversarial' broadly. We use these techniques to produce many paraphrases and find a subset of utterances that though very similar to the original test set, result in wrong predictions. We will measure the model robustness against such changes.
 
 Also note that to make the test set hard, we select only the examples for which the model prediction is different for the paraphrased sentence compared to the original sentence. We, however, do not use the original annotation for the perturbed sentences -- instead, we re-annotate the sentences manually. We explain the motivation and methodology for manual annotation later in this section.

\begin{table*}[hbt!]
\centering
\small
\begin{tabular}{lll}
  Original  & Variation 1  & Variation 2 \\
  \hline
 restart alarm for school& reboot the alarm for school (es)& reboot school alarm ($\bar{cs}$) \\
  are the roads icy & are there icy roads ($\bar{cs}$)& are the roads snowy (seq2seq)\\
   is the snow gone in seattle& the snows gone  (cs) &is the snow actually in seattle (seq2seq)   \\
    put all alarms on pause & put all alarms on a break (cs) & set all alarms on pause (es)\\
    pause my current alarm & suspend my current alarm (cs) &pausing my current alarm (es)
\end{tabular}
\caption{Adversarial examples alongside their original sentence. Note that we choose sentences on which the base model predicts intent differently than the original sentence.}
\label{tab:adversarial_examples}
\end{table*}

\subsection{Automatically Generating Examples}
We describe two methods of devising untargeted (not targeted towards a particular label) paraphrase generation to find a subset that dramatically reduce the accuracy of the model mentioned in the previous section. We follow \cite{luke_adversarial} and \cite{seq2seq_noise} to generate the potential set of sentences.

\subsubsection*{Back-translation}

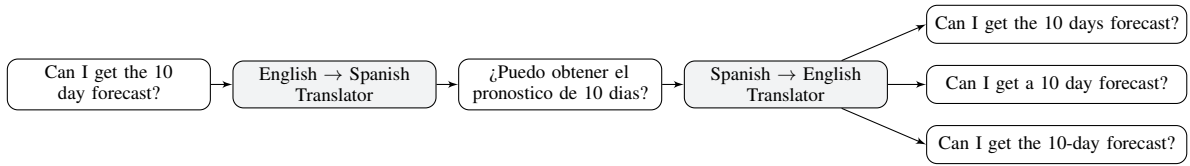
\begin{figure*}[t!]
\centering
\small
\makebox[0pt]{
    \begin{tikzpicture}[node distance = 1cm, scale=0.8, every node/.style={scale=0.8}, auto]
        % Place nodes
        \node [block3] (utterance) {Can I get the 10 day forecast?};
        \node [block2, right of=utterance, node distance=3.75cm] (enes) {English $\rightarrow$ Spanish\\Translator};
        \node [block3, right of=enes, node distance=3.75cm] (translation) {¿Puedo obtener el pronostico de 10 dias?};
        \node [block2, right of=translation, node distance=3.75cm] (esen) {Spanish $\rightarrow$ English\\Translator};
        \node [block4, right of=esen, node distance=4.5cm, yshift=1cm] (beam1) {Can I get the 10 days forecast?};
        \node [block4, right of=esen, node distance=4.5cm] (beam2) {Can I get a 10 day forecast?};
        \node [block4, right of=esen, node distance=4.5cm, yshift=-1cm] (beam3) {Can I get the 10-day forecast?};
        % Draw edges
        \path [line] (utterance) -- (enes);
        \path [line] (enes) -- (translation);
        \path [line] (translation) -- (esen);
        \path [line] (esen) -- (beam1.west);
        \path [line] (esen) -- (beam2.west);
        \path [line] (esen) -- (beam3.west);
        % \path [line] (components) -- node[text width=1.25cm, left=0.35cm, anchor=east] {Components} (trainer);
        % \path [line] (jobmanager) |- (init);
        % \path [line] (jobmanager) |- (components);
        % \path [line] (jobmanager) -- (trainer);
        % \path [line] (trainer) -- node[text width=1.25cm, text centered, above=.25cm, anchor=south] {PyTorch Model} (exporter);
        % \path [line] (exporter) -- (caffe2);
    \end{tikzpicture}
}
\caption{An example of back-translation. Translating the utterance back into the original language (English) but via an auxiliary language (Spanish) results in paraphrased variations in the beam.}
\label{fig:backtranslation}
\end{figure*}

Back-translation is a common technique in Machine Translation (MT) to improve translation performance, especially for low-resource language pairs~\cite{backtranslation_Sennrich,backtranslation-kenji, backtranslation_Hoang}. In back-translation, a MT system is used to translate the original sentences to an auxiliary language and a reverse MT system translates them back into the original language. At the final decoding phase, the top k beams are the variations of the original sentence. See Fig. \ref{fig:backtranslation}. \cite{luke_adversarial} which showed the effectiveness of simple back-translation in quickly generating adversarial paraphrases and breaking the correctly predicted examples.

To increase diversity, we use two different MT systems and two different auxiliary languages - Czech (cs) and Spanish (es), to use with our training data in English (en). We use the Nematus~\cite{nematus} pre-trained cs-en model, which was also used in~\cite{luke_adversarial}, as well as the FB internal MT system with pre-trained models for cs-en and es-en language pairs.  

\subsubsection*{Noisy Sequence Autoencoder}
Following~\cite{seq2seq_noise}, we train a sequence autoencoder~\cite{Dai2015seq} using all the training data. At test time, we add noise to the last hidden state of the encoder, which is used to decode a variation. We found that not using attention results in more diverse examples, by giving the model more freedom to stray from the original sentence. We again decode the top k beams as variations to the original sentence. We observed that the seq2seq model results in less meaningful sentences than using the MT systems, which have been trained over millions of sentences.

\subsection{Annotation}
For each of the above methods, we use the original test data and generate paraphrases using k=5 beams. We remove the beams that are the same as the original sentence after lower-casing. In order to make sure we have a high-quality adversarial test set, we need to manually check the model's prediction on the above automatically-generated datasets. Unlike targeted methods to procure adversarial examples, our datasets have been generated by random perturbations in the original sentences. Hence, we expect that the true adversarial examples would be quite sparse. In order to obviate the need for manual annotation of a large dataset to find these sparse examples, we sample only from the paraphrases for which the predicted intent is different from the original sentence's predicted intent. This significantly increases the chance of encountering an adversarial example. Note that the model accuracy on this test set might not be zero for two reasons: 1) the flipped intent might actually be justified and not a mistake. For example, \emph{``Cancel the alarm''} and \emph{``Pause the alarm''} may be considered as paraphrases, but in the dataset they correspond to alarm/cancel and alarm/pause intents, respectively, and 2) the model might have been making an error in the original prediction, which was corrected by the paraphrase. (However, upon manual observation, this rarely happens).

The other reason that we need manual annotation is that such unrestricted generation may result in new variations that can be meaningless or ambiguous without any context. Note that if the meaning can be easily inferred, we do not count slight grammatical errors as meaningless. Thus, we manually double annotate the sentences with flipped intent classification where the disagreements are resolved by a third annotator. As a part of this manual annotation, we also remove the meaningless and ambiguous sentences.  Note that these adversarial examples are untargeted, i.e., we had no control in which new label a perturbed example would be sent to.

\subsection{Analysis}
 We have shown adversarial examples from different sources alongside their original sentence in Table~\ref{tab:adversarial_examples}. We observe that some patterns, such as addition of a definite article or gerund appear more often in the es test set which perhaps stems from the properties of the Spanish language (i.e., most nouns have an article and present simple/continuous tense are often interchangeable). On the other hand, there is more verbal diversity in the cs test set which may be because of the linguistic distance of Czech from English compared with Spanish. Moreover, we observe many imperative-to-declarative transformation in all the back-translated examples. Finally, the seq2seq examples seem to have a higher degree of freedom but that can tip them off into the meaningless realm more often too. 
 
%  We also looked at what types of intent classification switches of the paraphrases are common. \mrinal{Review} Around $85\%$ of such intent prediction changes involve the ``unsupported`` intents, i.e., \textit{weather/unsupported} and \textit{alarm/unsupported}. These are sentences that do not belong to one of the supported labels the assistant does not know how to answer. We hypothesize that as such, sentences are very diverse, their class boundary is not smooth and hence, small perturbations can make the sentences cross that boundary easier than the boundaries corresponding to the more specific labels.

\begin{table*}[t!]
\centering
\small
\makebox[0pt]{
\begin{tabular}{|c|c||c|c|c|c||c|}
\hline
Model & Clean accuracy & Adv $\bar{cs}$ & Adv cs & Adv es & Adv seq2seq & Adv avg\\\hline
Baseline & 87.1  & 28.4  &34.2 &21.4 &32.8&29.2 \\\hline
Ensemble & \bf{88.2}& 31.8  &34.0 &23.7 &35.4& 31.2\\\hline
%es-self & 86.6 & 30.3 & 27.2 &18.2 &28.0 \\
Data augmentation (es) & 86.8 &  36.0 &	41.4	& 41.5 &37.5&39.1 \\
Data augmentation (cs) & 86.8& 37.0& \bf{50.7} &32.0 &36.5&39.0 \\
Data augmentation (cs+es) & 86.4&  37.1&47.4&40.8 &36.0&40.3 \\\hline
Clean logit & 84.9 & 26.5 &21.3 & 21.1& 32.3 & 25.3\\
Adv logit (es)  & 84.7&  36.4& 35.5 & \bf{45.3} &31.7 & 37.2\\
Adv logit + Data augmentation (es) & 83.8 &  39.4&  39.5 &41.6 & 32.8 & 38.3\\
Adv logit + Data augmentation (cs+es) & 83.6  & \bf{41.4}& 48.5  &42.5 &32.6 & \bf{41.3}\\
\makecell{Adv logit + Clean logit + \\Data augmentation (cs+es)} & 82.9 & 39.0& 42.7  &42.5 &\bf{38.7} & 40.7\\
\hline
\end{tabular}
}
\caption{Accuracy over clean and adversarial test sets. Note that data augmentation and logit pairing loss decrease accuracy on clean test sets and increase accuracy on the adversarial test sets.}
\label{tab:adversarial_FA}
\end{table*}

\section{Base Model}
A commonly used architecture for the task described in Section~\ref{sec:task} is a bidirectional LSTM for the sentence representation with separate projection layers for sentence (intent) classification and sequence word (slot) tagging~\cite{slotFillingYao,slotFillingMesnil,HakkaniTur2016,onenet}. 
In order to evaluate the model in a task oriented setting, exact match accuracy (from now on, accuracy) is of paramount importance. This is defined as the percentage of the sentences for which the intent and all the slots have been correctly tagged. 

We use two biLSTM layers of size $200$ and two feed-forward layers for the intents and the slots. We use dropout of $0.3$ and train the model for $20$ epochs with learning rate of $0.01$ and weight decay of $0.001$.  
This model, our baseline, achieves $87.1\%$ accuracy over the test set. 

The performance of the base model described in the previous section is shown in the first row of Table~\ref{tab:adversarial_FA} for the Nematus cs-en ($\bar{cs}$), FB MT system cs-en (cs) and es-en (es), sequence autoencoder (seq2seq), and the average of the adversarial sets (avg). We also included the results for the ensemble model, which combines the decisions of five separate baseline models that differ in batch order, initialization, and dropout masking. We can see that, similar to the case in computer vision~\cite{Goodfellow2014Explain}, the adversarial examples seem to stem from fundamental properties of the neural networks and ensembling helps only a little.

\section{Approaches to Improve Robustness}
In order to improve robustness of the base model against paraphrases and random noise, we propose two approaches: data augmentation and model smoothing via adversarial logit pairing. Data augmentation generates and adds training data without manual annotation. This would help the model see variations that it has not observed in the original training data. As discussed before, back-translation is one way to generate unlabeled data automatically. In this paper, we show how we can \textit{automatically} generate labels for such sentences during training time and show that it improves the robustness of the model. Note that for our task we have to automatically label both sentence labels (intent) and word tags (slots) for such sentences.

The second method we propose is adding logit pairing loss. Unlike data augmentation, logit pairing treats the original and paraphrased sentence sets differently. As such, in addition to the cross-entropy loss over the original training data, we would have another loss term enforcing that the predictions for a sentence and its paraphrases are similar in the logit space. This would ensure that the model makes smooth decisions and prevent the model from making drastically different decisions with small perturbations.

\subsection{Data Augmentation}
We generate back-translated data from the training data using  pre-trained FB MT system. We keep the top 5 beams after the back-translation and remove the beams that already exist in the training data after lower-casing. We observed that including the top 5 beams results in quite diverse combinations without hurting the readability of the sentences. In order to use the unlabeled data, we use an extended version of self training~\cite{self-training_mcclosky} in which the original classifier is used to annotate the unlabeled data. 
%One hopes that the benefit from better representation of the language model outweighs the noisy labels. 
Unsurprisingly, self-training can result in reinforcing the model errors. Since the sentence intents usually remain the same after paraphrasing for each paraphrase, we annotate its intent as the intent of the original sentence. 
Since many slot texts may be altered or removed during back-translation,
%using the same method for the slots (via following the translation weights as described in~\cite{sebastianMultilingual}) did not work for us). 
we use self-training to label the slots of the paraphrases. We train the model on the combined clean and noisy datasets with the loss function being the original loss plus the loss on back-translated data weighted by 0.1 for which the accuracy on the clean dev set is still negligible. The model seemed to be quite insensitive against this weight, though and the clean dev accuracy was hurt by less than 1 point using weighing the augmented data equally as the original data. The accuracy over the clean test set using the augmented training data having Czech (cs) and Spanish (es) as the auxiliary languages are shown in~Table~\ref{tab:adversarial_FA}. 

We observe that, as expected, data augmentation improves accuracy on sentences generated using back-translation, however we see that it also improves accuracy on sentences generated using seq2seq autoencoder. We discuss the results in more detail in the next section.

\subsection{Model smoothing via Logit Pairing}
% Perturbing the input image to a classifier and adding a loss to enforce the new and original outputs to be the same has been shown to be an effective way of smoothing the classifier decision boundaries and making it more resistant toward the adversarial input~\cite{Goodfellow2014Explain,Miyato_VAT}.

\cite{logit_pairing_2018} perturb images with the attacks introduced by~\cite{madry2018towards} and report state-of-the-art results by matching the logit distribution of the perturbed and original images instead of matching only the classifier decision. They also introduce clean pairing in which the logit pairing is applied to random data points in the clean training data, which yields surprisingly good results.
%which is because the prevention of the model from being too certain in its decision making.
Here, we modify both methods for the language understanding task, including sequence word tagging, and expand the approach to targeted pairing for increasing robustness against adversarial examples.

\subsubsection{Clean Logit Pairing}
Pairing random queries as proposed by~\cite{logit_pairing_2018} performed very poorly on our task. In the paper, we study the effects when we pair the sentences that have the same annotations, i.e., same intent and same slot labels. Consider a batch $M$, with $m$ clean sentences. For each tuple of intent and slot labels, we identify corresponding sentences in the batch, $M_k$ and sample pairs of sentences. We add a second cost function to the original cost function for the batch that enforces the logit vectors of the intent and same-label slots of those pairs of sentences to have similar distributions:
\begin{equation*}
    \frac{\lambda_{sf}}{P} \sum_{i,j \in M_k}\left( L(I^{(i)},I^{(j)})+ \sum_s L(S^{(i)}_{s},S^{(j)}_{s})\right)
\end{equation*}
 where $I^{(i)}$ and $S^{(i)}_s$ denote the logit vectors corresponding to the intent and $s^{th}$ slot of the $i^{th}$ sentence, respectively. Moreover, $P$ is the total number of sampled pairs, and $\lambda_{sf}$ is a hyper-parameter. We sum the above loss for all the unique tuples of labels and normalize by the total number of pairs. Throughout this section, we use MSE loss for the function $L()$. We train the model with the same parameters as in Section~\ref{sec:task}, with the only difference being that we use learning rate of $0.001$ and train for $25$ epochs to improve model convergence. Contrary to what we expected, clean logit pairing on its own reduces accuracy on both clean and adversarial test sets. Our hypothesis is that the logit smoothing resulted by this method prevents the model from using weakly correlated features~\cite{robustness2018odds}, which could have helped the accuracy of both the clean and adversarial test sets. 
%  We, however, discuss in the appendix that the approach might be useful for more `consistent' predictions.
 
\subsubsection{Adversarial Logit Pairing (ALP)}
In order to make the model more robust to paraphrases, we pair a sentence with its back-translated paraphrases and impose the logit distributions to be similar. We generate the paraphrases using the FB MT system as in the previous section using es and cs as auxiliary languages. For the sentences $m^{(i)}$ inside the mini-batch and their paraphrase $\tilde{m}^{(i)}_k$, we add the following loss 
\vspace{-10pt}
\begin{equation*}
  \frac{\lambda_{a}}{P} \sum_{k,i}\left(   L(I^{(i)},\tilde{I}^{(i)}_k) + \sum_s L(S^{(i)},\tilde{S}^{(i)}_k) \right)
  \label{eq:adversarial_logit}
\end{equation*}
where $P$ is the total number of original-paraphrase sentence pairs.
Note that the first term, which pairs the logit vectors of the predicted intents of a sentence and its paraphrase, can be obtained in an unsupervised fashion. For the second term however, we need to know the position of the slots in the paraphrases in order to be matched with the original slots. We use self-training again to tag the slots in the paraphrased sentence. Then, we pair the logit vectors corresponding to the common labels found among the original and paraphrases slots left to right. We also find that adding a similar loss for pairs of paraphrases of the original sentence, i.e. matching the logit vectors corresponding to the intent and slots, can help the performance on the accuracy over the adversarial test sets.  In Table~\ref{tab:adversarial_FA}, we show the results using ALP (using both the original-paraphrase and paraphrase-paraphrase pairs) for $\lambda_a=0.01$.

\section{Results and Discussion}

We observe that data augmentation using back-translation improves the accuracy across all the adversarial sets, including the seq2seq test set.  Unsurprisingly, the gains are the highest when augmenting the training data using the same MT system and the same auxiliary language that the adversarial test set was generated from. However, more interestingly, it is still effective for adversarial examples generated using a different auxiliary language or a different MT system (which, as discussed in the previous section, yielded different types of sentences) from that which was used at the training time. More importantly, even if the generation process is different altogether, that is, the seq2seq dataset generated by the noisy autoencoder, some of the gains are still transferred and the accuracy over the adversarial examples increases.  We also train a model using the es and cs back-translated data combined. Table~\ref{tab:adversarial_FA} shows that this improves the average performance over the adversarial sets.

This suggests that in order to achieve robustness towards different types of paraphrasing, we would need to augment the training data using data generated with various techniques. But one can hope that some of the defense would be transferred for adversarial examples that come from unknown sources. Note that unlike the manually annotated test sets, the augmented training data contains noise both in the generation step (e.g. meaningless utterances) as well as in the automatic annotation step. But the model seems to be quite robust toward this random noise; its accuracy over the clean test set is almost unchanged while yielding nontrivial gains over the adversarial test sets.

We observe that ALP results in similarly competitive performance on the adversarial test sets as using the data augmentation but it has a more detrimental effect on the clean test set accuracy. We hypothesize that data augmentation helps with smoothing the decision boundaries without preventing the model from using weakly correlated features. Hence, the regression on the clean test set is very small. This is in contrast with adversarial defense mechanisms such as ALP~\cite{robustness2018odds} which makes the model regress much more on the clean test set.

We also combine ALP with the data augmentation technique that yields the highest accuracy on the adversarial test sets but incurs additional costs to the clean test set (more than three points compared with the base model). Adding clean logit pairing to the above resulted in the most defense transfer (i.e. accuracy on the seq2seq adversarial test set) but it is detrimental to almost all the other metrics. One possible explanation can be that the additional regularization stemming from the clean logit pairing helps with generalization (and hence, the transfer) from the back-translated augmented data to the seq2seq test set but it is not helpful otherwise.

\section{Related Work}

Adversarial examples~\cite{Goodfellow2014Explain} refer to intentionally devised inputs by an adversary which causes the model's accuracy to make highly-confident but erroneous predictions, e.g. Fast Gradient Sign Attack (FGSA) ~\cite{Goodfellow2014Explain} and Projected gradient Descent (PGD) \cite{madry2018towards}. In such methods, the constrained perturbation that (approximately) maximizes the loss for an original data point is added to it. In white-box attacks, the perturbations are chosen to maximize the model loss for the original inputs~\cite{Goodfellow2014Explain, madry2018towards, Papernot2016Limitation}. Such attacks have shown to be transferable to other models which makes it possible to devise black-box attacks for a machine learning model by transferring from a known model~\cite{papernot_transfer,Papernot2017Blackbox}.

Defense against such examples has been an elusive task, with proposed mechanisms proving effective against only particular attacks ~\cite{madry2018towards,raghunathan2018certified}. Adversarial training~\cite{Goodfellow2014Explain} augments the training data with carefully picked perturbations during the training time, which is robust against normed-ball perturbations. But in the general setting of having unrestricted adversarial examples, these defenses have been shown to be highly ineffective~\cite{adversarial_gan}.

\cite{Ebrahimi2017HotFlipWA} introduced white-box attacks for language by swapping one token for another based on the gradient of the input. \cite{generate2018alzantot} introduced an algorithm to generate adversarial examples for sentiment analysis and textual entailment by replacing words of the sentence with similar tokens that preserve the language model scoring and maximize the target class probability. ~\cite{Miyato_VAT_NLP} introduced one of the few defense mechanisms for NLP by extending adversarial training to this domain by perturbing the input embeddings and enforcing the label (distribution) to remain unchanged. ~\cite{adversarial_relation_extraction} and ~\cite{adversarial_pos} used this strategy as a regularization method for part-of-speech, relation extraction and NER tasks. Such perturbations resemble the normed-ball attacks for images but the perturbed input does not correspond to a real adversarial example. \cite{luke_adversarial} studied two methods of generating adversarial data -- back-translation and syntax-controlled sequence-to-sequence generation.
They show that although the latter method is more effective in generating syntactically diverse examples, the former is also a fast and effective way of generating adversarial examples.

There has been a large body of literature on language understanding for task oriented dialog using the intent/slot framework. Bidirectional LSTM for the sentence representation alongside separate projection layers for intent and slot tagging is the typical architecture for the joint task~\cite{slotFillingYao,slotFillingMesnil,HakkaniTur2016,onenet}. 
 
In parallel to the current work,~\cite{Quoc_augmentation} introduced unsupervised data augmentation for classification tasks by perturbing the training data and similar to \cite{Miyato_VAT_NLP} minimize the KL divergence between the predicted  distributions on an unlabeled example and its perturbations. Their goal is to achieve high accuracy using as little labeled data as possible by leveraging the unlabeled data. In this paper, we have focused on increasing the model performance on adversarial test sets in supervised settings while constraining the degradation on the clean test set. Moreover, we focused on a more complicated task: the joint classification and sequence tagging task.

\section{Conclusion}

In this paper, we study the robustness of language understanding models for the joint task of sentence classification and sequence word tagging in the field of task oriented dialog by generating adversarial test sets. We further discuss defense mechanisms using data augmentation and adversarial logit pairing loss. 

We first generate adversarial test sets using two methods, back-translation with two languages and sequence auto-encoder, and observe that the two methods generate different types of sentences. Our experiments show that creating the test set using a combination of the two methods above is better than either method alone, based on the model's performance on the test sets. Secondly, we propose how to improve the model's robustness against such adversarial test sets by both augmenting the training data and using a new loss function based on logit pairing with back-translated paraphrases annotated using self-training. The experiments show that combining data augmentation using back-translation and adversarial logit pairing loss performs best on the adversarial test sets. 

 \subsubsection*{Future Work} Though the adversarial accuracy has significantly improved using the above techniques, there is still a huge gap between the adversarial and clean test accuracy. Exploring other techniques to augment the data as well as other methods to leverage the augmented data is left for future work. For example, using sampling at the decoding time~\cite{backtranslation-kenji} or conditioning the seq2seq model on structure~\cite{luke_adversarial} has shown to produce more diverse examples. On the other hand, using more novel techniques such as multi-task tri-training~\cite{ruder_semisupervised} to label the unlabeled data rather than the simple self-training may yield better performance. Moreover, augmenting the whole dataset through back-translation and keeping the top k beams is not practical for large datasets. Exploring more efficient augmentations, i.e., which sentences to augment and which back-translated beams to keep, and adapting techniques such as in ~\cite{kuchnik2018subsampling} are also interesting research directions to pursue. 

In this paper, we studied various ways of devising untargeted adversarial examples. This is in contrast with targeted attacks, which can perturb the original input data toward a particular label class. Encoding this information in the seq2seq model, e.g., feeding a one-hot encoded label vector, may deserve attention for future research. 

\small
\bibliographystyle{plain}
\bibliography{neurips_2019}
\end{document}